\pdfoutput=1

\documentclass[11pt]{article}

\usepackage{PRIMEarxiv}

\usepackage{times}
\usepackage{latexsym}
\usepackage{url}

\usepackage[T1]{fontenc}

\usepackage[utf8]{inputenc}

\usepackage{microtype}

\usepackage{inconsolata}

\usepackage{graphicx}

\title{StructuredRAG: JSON Response Formatting with Large Language Models}

\author{
    Connor Shorten \\
    Weaviate \\
    \And
    Charles Pierse \\
    Weaviate \\
    \And
    Thomas Benjamin Smith \\
    Weaviate \\
    \And
    Erika Cardenas \\
    Weaviate \\
    \And
    Akanksha Sharma \\
    Weaviate \\
    \And
    John Trengrove \\
    Weaviate \\
    \And
    Bob van Luijt \\
    Weaviate \\
}

\begin{document}
\maketitle
\begin{abstract}

The ability of Large Language Models (LLMs) to generate structured outputs, such as JSON, is crucial for their use in Compound AI Systems. However, evaluating and improving this capability remains challenging. In this work, we introduce StructuredRAG, a benchmark of six tasks designed to assess LLMs’ proficiency in following response format instructions. We evaluate two state-of-the-art LLMs, Gemini 1.5 Pro and Llama 3 8B-instruct with 4-bit quantization using two distinct prompting strategies. We introduce these prompting strategies as f-String and Follow the Format (FF) prompting. Across 24 experiments, we find an average success rate of 82.55\%. We further find a high variance in performance across tasks, models, and prompting strategies with success rates ranging from 0 to 100\%. We find that Llama 3 8B-instruct often performs competitively with Gemini 1.5 Pro. We observe that task complexity significantly influences performance, with tasks involving lists or composite object outputs proving more challenging. Our findings highlight the need for further research into improving the reliability and consistency of structured output generation in LLMs. We have open-sourced our experimental code and results at github.com/weaviate/structured-rag.

\end{abstract}

\section{Introduction}

Large Language Models (LLMs) have become extremely effective at Zero-Shot Learning. Zero-Shot Learning is used to describe a machine learning model’s ability to perform a task without any training data for the task given in advance. An emergent area of importance is not only to test LLMs on how well they can perform novel tasks, but also how well they can structure their output in a particular format. This is a critical requirement for developing Compound AI Systems \cite{1, 3} that consist of multiple LLM inferences or external computational tools. For example, Multi-Hop RAG \cite{2} is a Compound AI System where an LLM inference first predicts one or multiple search queries for an input and then sends these queries to a search tool. Another LLM inference then aggregates these search results and the original question to generate a response. In order for the Multi-Hop RAG system to parse the response from the query writer to send to the search tool, it is critical that the query writer follows a particular response format such as a JSON with the key “queries” and a list of strings as the value.

\begin{figure*}
    \centering
    \includegraphics[width=1\linewidth]{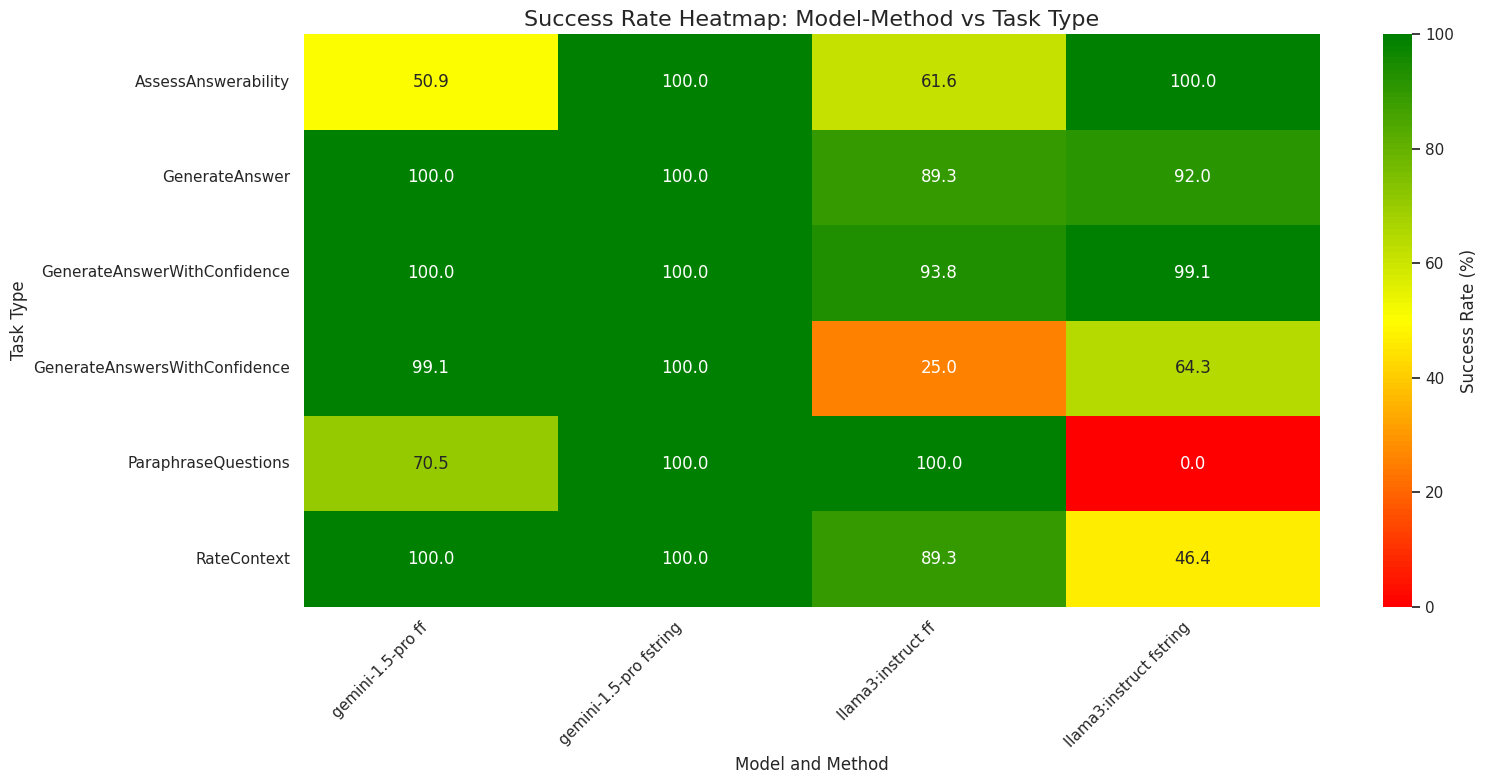}
    \caption{An overview of our experimental results. Across 24 experiments, we achieve an average response format success rate of 82.55\%. However, we find high variance in these results, 11 out of the 24 tests achieve 100\% success, 2 out of 24 achieve 25\% success or lower, and 5 of the tested methods achieve between 45\% to 75\% success..}
    \label{fig:enter-label}
\end{figure*}

In this work, we seek to measure the ability of LLMs to follow JSON response format instructions with Zero-Shot Learning. While structured decoding methods, such as DOMINO \cite{16}, have emerged as a popular solution for ensuring correct JSON outputs in Compound AI Systems, we seek to better understand the baseline performance of Zero-Shot Learning. Structured decoding may slow down inference throughput, complicate system integration, and interfere with the LLM's prior knowledge and the benefits of prompt optimization \cite{LetMeSpeak}. To address these concerns, we a construct a novel benchmark of six RAG-inspired \cite{7} structured output tests. These tests explore different typed JSON responses such as string, integer, or boolean values, as well as outputting a list of strings, denoted as List[string]. Further, we illustrate the use of composite objects containing more than one type per instance. We present the AnswerWithConfidence composite object consisting of a string valued answer and an integer valued confidence. We further test the ability to output a list of AnswerWithConfidence objects, similarly denoted as List[AnswerWithConfidence]. An output from the LLM passes these tests if it is able to be parsed into the requested JSON response format. This entails that the output jointly contains the correct keys as well as the correct types per value. We run our tests using the Gemini 1.5 Pro API \cite{4} and Llama 3 8B-instruct with 4-bit quantization \cite{5} hosted with Ollama \cite{Ollama}, restricting the task to avoid the use of structured decoding methods, which we further discuss later in our paper. In this study, we measure the success rate of parsing LLM responses into the desired JSON format. We leave inference throughput comparisons and the entanglement of response formatting and task performance, such as answer correctness, for future work.

Figure 1 presents an overview of our results. We find much better performance when tasked with simple output types such as a single string, integer, or boolean value, whereas performance degrades significantly on list outputs and composite objects. We find two cases of high failure rates from Llama 3 8B-instruct when outputting a list of strings in ParaphraseQuestions and when outputting a list of composite objects in GenerateAnswersWithConfidences. Out of the 24 experimental trials, we find 11 cases where the structured output succeeds 100\% of the time. We find that both Gemini 1.5 Pro and Llama 3 8B-instruct show comparable performance on this benchmark, with each model excelling in different tasks. Our results indicate the need for future research on generating structured outputs.

Our contributions are as follows:
\begin{itemize}
    \item We introduce StructuredRAG, a set of six structured output tests that can be adapted to Retrieval-Augmented Generation systems consisting of questions and supplemental context.
    \item We compare f-String with Follow the Format (FF) prompting across the Gemini 1.5 Pro and Llama 3 8B-instruct LLMs. We find a high variance in success rates across tested methods. Gemini 1.5 Pro outperforms Llama 3 8B-instruct achieving an average success rate of 93.4\% compared to 71.7\%. We do not find a significant difference in success rates between f-String and FF prompting. We find a performance gap from single structured outputs to more complex outputs such as lists, composite objects, and lists of composite objects.
    \item We demonstrate the effectiveness of OPRO prompt optimization on JSON response formatting with Llama 3 8B-instruct, achieving a 100\% success rate on the task of outputting a list of composite objects.
\end{itemize}

\section{Methodology}

We present WikiQuestions, a dataset that isolates the generation aspect of Retrieval-Augmented Generation systems \cite{7}. WikiQuestions contains 56 randomly selected Wikipedia titles and abstracts. Each Wikipedia title-abstract example contains an answerable question and an unanswerable question generated with Gemini Pro 1.5 \cite{4} and further verified with human supervision. Thus we end up with 112 questions for each test. An example from our WikiQuestions dataset is shown in Figure 2. At 30 tokens, the abstract shown is the shortest example in our dataset. We further experiment with longer abstracts up to 500 tokens. We believe it will be an interesting opportunity for future work to explore how performance on the StructuredRAG benchmark scales when dealing with longer or noisy contexts retrieved from a search database. We hypothesize that this will make the task more challenging based on previous work from Liu et al. \cite{8} and Shi et al. \cite{9}.

\begin{figure}
    \centering
    \includegraphics[scale=0.5]{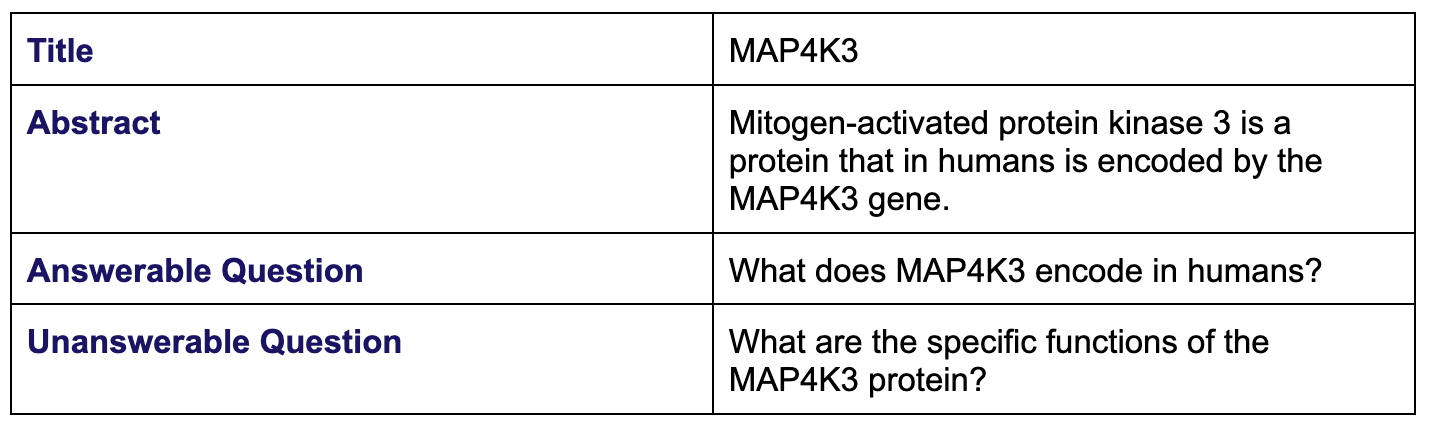}
    \caption{An illustration of the WikiQuestions dataset. Title-Abstract pairs are randomly sampled from Wikipedia. Gemini 1.5 Pro then synthesizes an answerable and unanswerable question for each. These generated questions are validated by a human annotator.}
    \label{fig:enter-label}
\end{figure}

We present the Structured RAG benchmark, consisting of six tests to measure response format instruction following. Structured RAG tests response formats for string, integer, boolean, list of strings, and a composite AnswerWithConfidence object, as well as a list of AnswerWithConfidence objects. The most common failure case is for the LLM API to respond by acknowledging the task with a response such as, "Sure, I can help you with that!”, or, “Here is the output in the required JSON format:”. Interestingly, our Follow the Format (FF) prompting method inspired by DSPy, provokes the model to occasionally fail by adding a “Reasoning: …” follow up to its generation, akin to how Chain-of-Thought prompting \cite{10} is implemented in DSPy \cite{21, 22}. We further consider cases where the model outputs a string type instead of an integer as a failure. For example, an output of \{"context\_score": “4”\} is considered a failure. The six tasks performed, their output type, and a successful example are illustrated in Figure 6. The StructuredRAG benchmark presents structured type outputs aligned with Retrieval-Augmented Generation systems. The StructuredRAG tests are not specific to WikiQuestions and can be adapted to any RAG system.

\subsection{Model Comparison}

We compare Gemini 1.5 Pro \cite{4} with a 4-bit quantized version of an instruction tuned variant of Llama 3 8B \cite{5, Ollama}. We chose these models to better understand the performance gap between the largest, and generally most capable language models available, such as Gemini 1.5 Pro, with smaller models, such as Llama 3 8B-instruct. We find that Llama 3 8B-instruct performs competitively with Gemini 1.5 Pro when comparing average performance across all tasks. However, as shown in Figure 4, we find a high variance in Llama 3 8B-instruct performance due to massive failures on two tasks. All tests are run single-threaded single-node. All LLM APIs are tested with a temperature set to 0 to minimize randomness in the responses. We leave it to future work to explore the performance of additional LLMs.

\begin{figure*}
    \centering
    \includegraphics[width=0.8\linewidth]{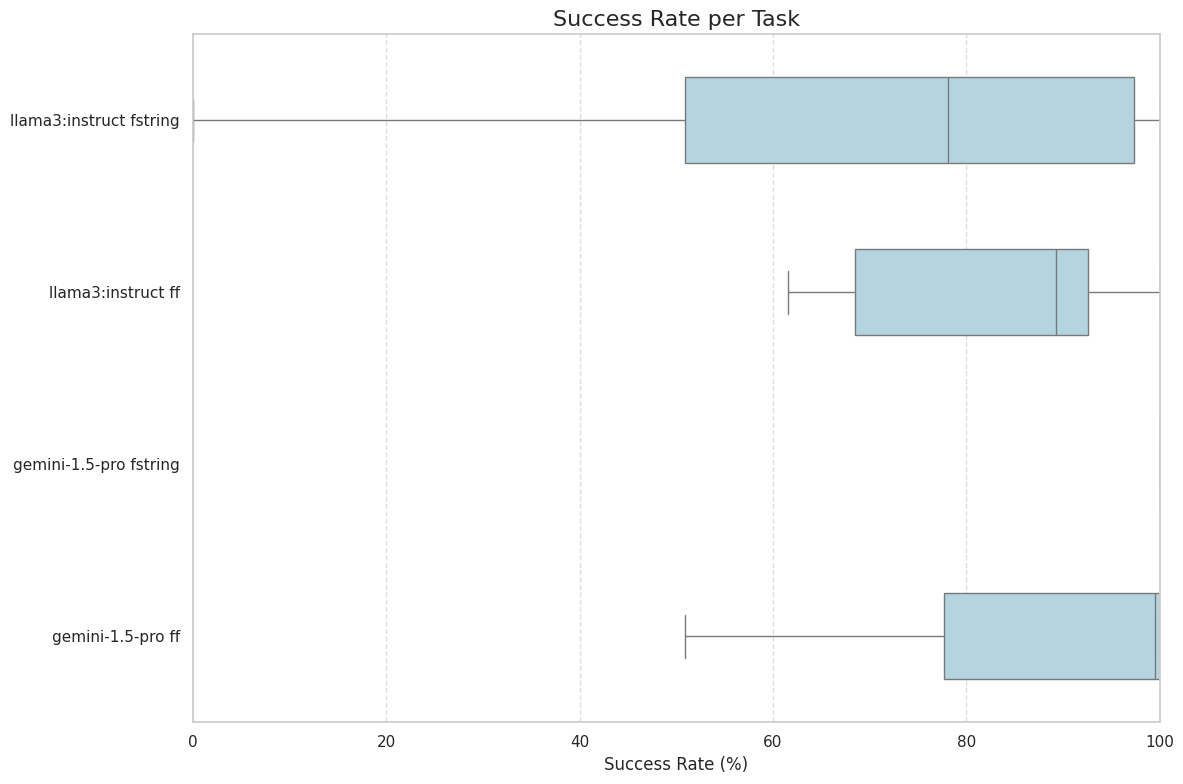}
    \caption{A visualization of performance variance across all tasks from each model and prompting strategy tested.}
    \label{fig:enter-label}
\end{figure*}

\subsection{Prompting Strategies}

We compare two prompting strategies for response formatting, which we introduce as f-String prompting and Follow the Format (FF) prompting. We define, f-String prompting to describe embedding the task-specific variables within the prompt, whereas Follow the Format (FF) prompting follows a more rigid format of first explaining the task, then the response format, and then the input-specific variables. The task\_instructions and response\_format input variables are consistent across the StructuredRAG tasks, such as ParaphraseQuestions. For example, the ParaphraseQuestions task has the task\_instructions, “Generate 3 paraphrased versions of the given question.” and the "response\_format”: “\{“answer”: “string”, “confidence”: “int (0-5)”\}”. The references then inject a particular context and question pair for the inference. An example of f-String and FF prompting is shown in Figure 7. FF prompting, as used in the DSPy framework, clearly distinguishes between general instructions, formatting requirements, and task-specific inputs. Our study aims to understand if the more natural language style of f-String prompting offers a performacne advantage over FF prompting. We do not find such a benefit. While we achieve a successful result by applying the OPRO optimizer to the FF prompting approach, further research is needed to understand the interplay between priors in prompting structures and emerging optimization methods.

\subsection{Summary of Zero-Shot Results}

The results of our experiments on the StructuredRAG benchmark reveal several key findings. Shown in Figures 1 and 4, we see a very high variance in performance. We find cases of poor performance, particularly when testing Llama 3 8B-instruct with f-String prompting on Paraphrase Questions (0\% success rate) and when testing Llama 3 8B-instruct with FF prompting on GenerateAnswersWithConfidence (25\% success rate). Gemini 1.5 Pro outperforms Llama 3 8B-instruct achieving an average success rate of 93.4\% compared to 71.7\%. We find lower success rates with ParaphraseQuestions, involving a list of strings output, and GenerateAnswersWithConfidences, requiring a list of composite objects as output, achieving average success rates of 72.1\% and 67.6\% respectively. These tasks are also where the gap between Gemini 1.5 Pro and Llama 3 8B-instruct is further pronounced with Gemini 1.5 Pro achieving success rates of 99.6\% and 85.3\% compared to Llama 3 8B-instruct's scores of 44.7\% and 50\%. We find f-String prompting to outperform FF prompting for Gemini 1.5 Pro, achieving a superior 100\% success rate compared to 86.8\%. Alternately, we find FF prompting to perform better for Llama 3 8B-instruct, achieving 76.5\% success compared to 67.0\%.

\begin{figure*}
    \centering
    \includegraphics[width=1\linewidth]{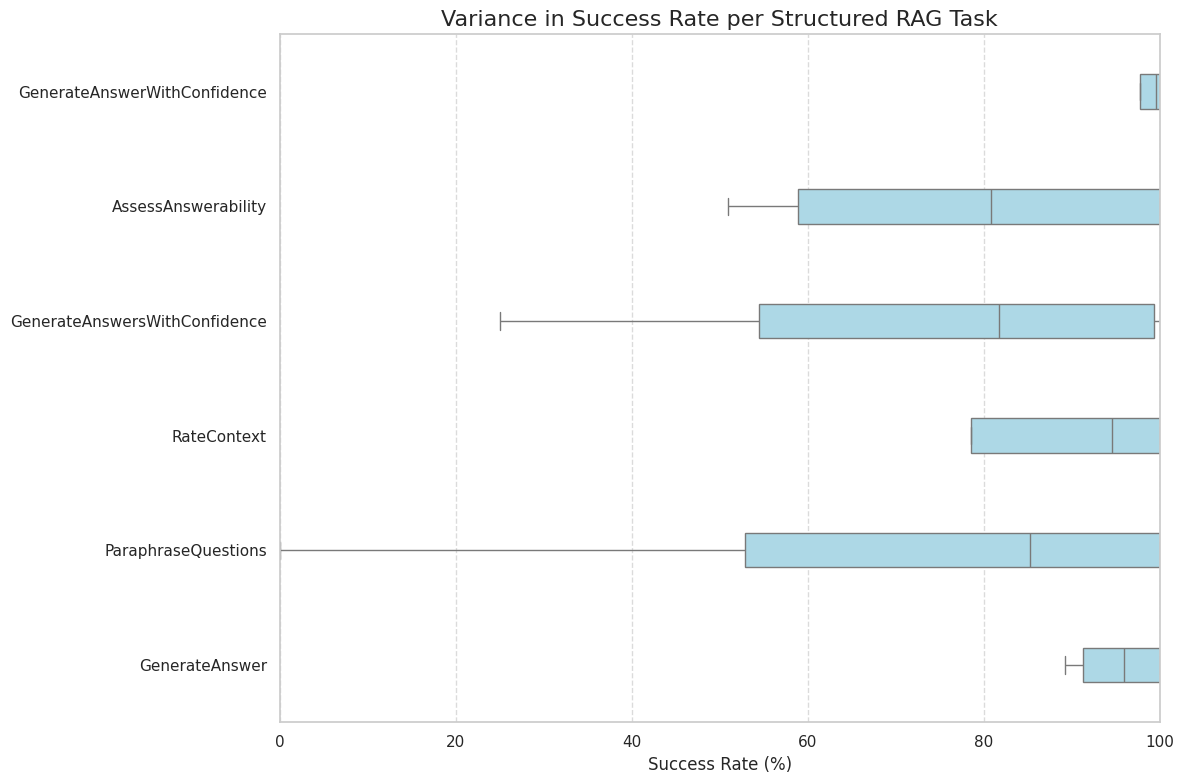}
    \caption{An illustration of performance variance across StructuredRAG tasks.}
    \label{fig:enter-label}
\end{figure*}

\subsection{OPRO JSON Response Optimization}

We then test the OPRO optimizer \cite{25} on optimizing the Llama 3 8B-instruct prompt for the GenerateAnswersWithConfidences task, requiring outputting a list of composite objects. OPRO prompt optimization works by firstly proposing tweaks to the original prompt and scoring the new prompts with a metric, in our case JSON validation, and a training set, of which we divide our WikiQuestions dataset into 40 training and 16 testing examples. OPRO then uses the information from how each candidate prompt scored with respect to the metric to propose new prompts. We use the GPT-4 LLM \cite{GPT4} to propose paraphrasings. We set OPRO to propose 25 new candidates per round and continue for 2 rounds. Although, we note OPRO finds the prompt that succeeds 100\% of the time within the first 10 paraphrasings of the first round. The final prompt discovered by OPRO optimization is shown in Figure 7. The optimized prompt adds notes such as, "Review the task\_instructions meticulously, ensuring thorough comprehension before beginning your response", as well as several other instructions, "to guarantee a pure and correct JSON output". The success of OPRO optimization suggests a path towards JSON response formatting without additional decoding methods. We leave it to future work to understand the impact of more advanced prompt optimizers such as MIPRO \cite{24} or BetterTogether \cite{11} on StructuredRAG.

\section{Discussion}

\subsection{Untested Solutions}

In this work, we aim to present the StructuredRAG benchmark and a few simple baselines and their resulting performance. We note that there are several promising directions to improve performance on this test such as ensembling, retry mechanisms, and chain-of-thought prompting. Ensembling describes leveraging the stochastic nature of LLMs to produce multiple outputs per input \cite{12, 13}. In our experiments, we average the results across 3 trials. We found consistent results, suggesting that trying again naively would be unlikely to succeed. However, we do find significant variance between f-String and FF prompting on some tasks, suggesting that ensembling with prompt paraphrasings could be promising.

It may additionally be promising to test Chain-of-Thought (CoT) prompting strategies \cite{10}. This entails adding a “rationale” key to the model’s output such that the additional reasoning improves the performance of the model. However, this will require the output to be a composite object with the additional "rationale" key and string-valued response. Our results suggest that this additional output structure may result in lower success rates. Similar in spirit to structured decoding methods, it may also be helpful to prefix the end of the prompt with “\{“ or use the key, such as “\{“paraphrased\_questions”: [".

Another promising approach is to apply a “retry” prompt on failed outputs. For example, an f-string prompt could be: “A system has produced the output: \{output\}. This output has been judged to have failed the response format instructions given here: \{response\_format\}. Please correct the output to the desired response format”. This is similar to Reflexion \cite{14} prompting that introduces the idea of chaining LLM calls with self-reflection. This paradigm of LLM computing is also being pioneered by works such as DSPy Assertions \cite{15} or SPADE \cite{SPADE} that further tackle verifying and correcting outputs in Compound AI Systems \cite{NoNs}.

 \section{Related Work}

\subsection{Structured Decoding}

Our work on StructuredRAG is closely related to recent advancements in Structured Decoding with Large Language Models (LLMs). The Language Model Query Language (LMQL) \cite{LMQL} presents a query language to combine prompting with output constraints and structure. The DOMINO algorithm \cite{16} further presents a novel experimentation of constrained decoding methods for LLMs. DOMINO is similarly evaluated on JSON generation, as well as Mermaid flowchart creation and function call generation. We note that function call generation is very similar to the task of following JSON response format instructions. However, we believe a critical distinction is that function call generation works are further concerned with routing a task to particular functions \cite{19}. DOMINO leverages many advanced Structured Decoding methods such as pre-computation, speculative decoding \cite{20}, and opportunistic masking. DOMINO measures both response format accuracy as well as downstream task accuracy, which we leave to future work. DOMINO achieves up to 1.77x throughput improvement over unconstrained generation for JSON tasks and improves performance from 37.6\% to 98.8\% in the newly introduced QuizGen task.

This work is related to Formal Grammars in Large Language Model sampling. Grammar-constrained decoding (GCD), led by Geng et al. \cite{17} presents formal grammars for information extraction. JSON is similarly described by a context-free grammar which defines its structure including objects, arrays, strings, numbers, and boolean values. GCD similarly targets constrained decoding without fine-tuning model weights, aiming to generate structured outputs while maintaining the model’s general capabilities and flexibility. Wang presents YieldLang \cite{18}, introducing key system engineering concepts such as asynchronous parsing. YieldLang proposes a coroutine-based framework for generating domain-specific languages (DSLs) using Python’s yield keyword. This achieves efficient parsing and generation of DSLs, akin to context-free grammar generation. These works on Structured Decoding primarily focus on efficient execution and highlight the growing importance of structured outputs in AI systems, which leads us to consider their role in Compound AI Systems.

\subsection{Compound AI Systems}

Our work addresses a crucial need in the development of Compound AI Systems. From Zaharia et al. \cite{1}, “state-of-the-art AI results are increasingly obtained by compound systems with multiple components, not just monolithic models”. These components typically require a particular output format in order to be parsed and sent to the next component. Further explored in works such as ALTO from Santhanam et al. \cite{3}, there are many emerging optimizations available for engineering Compound AI Systems. ALTO illustrates how structured response formats enable streaming intermediate responses, with further investigation into systems such as an aggregation-aware routing interface and distributed prompt-aware scheduling. Zheng et al. have further introduced SGLang, "a comprehensive system for efficient execution of complex language model programs" \cite{23}.  SGLang addresses challenges in programming and executing LM programs that require multiple generation calls, advanced prompting techniques, and structured inputs/outputs with key innovations include RadixAttention for efficient KV cache reuse, compressed finite state machines for faster constrained decoding, and API speculative execution for API-only models. We leave it to future work to measure SGLang and vLLM \cite{vLLM} style decoding and inference throughput on benchmarks such as StructuredRAG.

\section{Conclusion}

In conclusion, this study introduces the StructuredRAG benchmark for assessing Large Language Models' ability to generate structured outputs without relying on structured decoding methods. Our experiments with Gemini 1.5 Pro and Llama 3 8B-instruct reveal significant variability in performance across different structured generation tasks and prompting strategies. The results highlight the varying capabilities of LLMs in generating complex structured outputs. While some models struggle with lists and composite objects, others show high performance on these tasks, indicating the potential for further improvements. We leave it to future work to explore advanced techniques such as ensembling, retry mechanisms, chain-of-thought prompting and prompt optimization to further enhance performance on response formatting without structured decoding methods.

\bibliographystyle{unsrt}  
\bibliography{references.bib}

\appendix

\begin{figure*}
    \centering
    \includegraphics[width=1\linewidth]{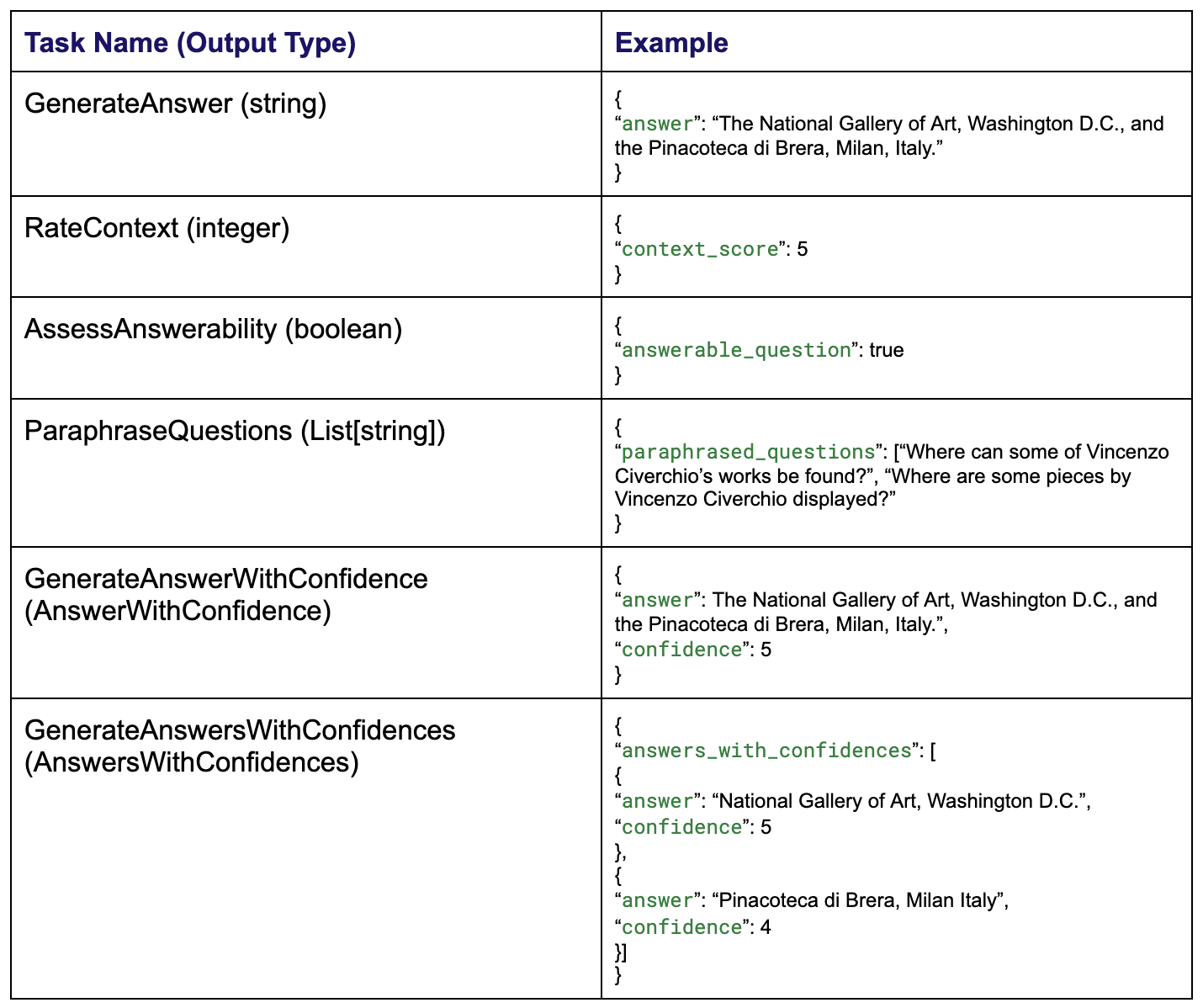}
    \caption{An overview of the StructuredRAG benchmark. StructuredRAG tests response formatting across six different output type tests, string, integer, boolean, List[string], AnswerWithConfidence, and List[AnswerWithConfidence].}
    \label{fig:enter-label}
\end{figure*}

\begin{figure*}
    \centering
    \includegraphics[width=1\linewidth]{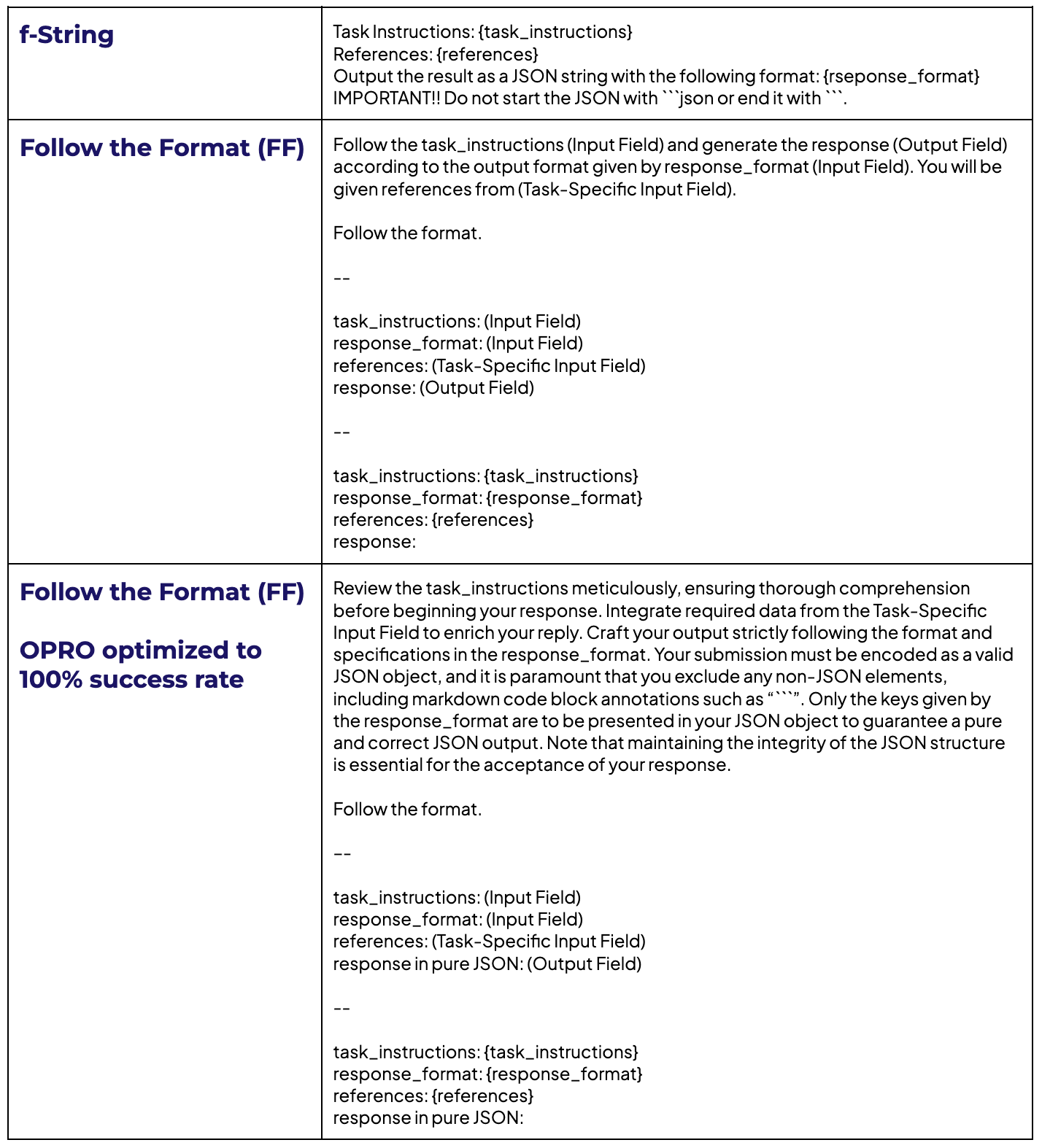}
    \caption{An illustration of the distinction between f-String and Follow the Format (FF) prompting strategies.}
    \label{fig:enter-label}
\end{figure*}

\end{document}